\documentclass[runningheads]{llncs}
\usepackage{graphicx}
\usepackage{multirow}
\usepackage{verbatim}
\usepackage{array}
\usepackage{pgfplots}
\usepackage{wrapfig}
\usepackage{booktabs}
\usepackage{amsmath}
\usepackage{marvosym}

\usepackage[hidelinks,hypertexnames=false]{hyperref}

\newcolumntype{?}{!{\vrule width 2pt}}
\makeatletter
\newcommand{\thickhline}{%
    \noalign {\ifnum 0=`}\fi \hrule height 2pt
    \futurelet \reserved@a \@xhline
}
\newcolumntype{"}{@{\hskip\tabcolsep\vrule width 2pt\hskip\tabcolsep}}
\makeatother

\begin{document}

\title{Graph Neural Network Approach to Semantic Type Detection in Tables \thanks{Published at Pacific-Asia Conference on Knowledge Discovery and Data Mining 2024}}

\author{Ehsan Hoseinzade\thanks{Corresponding author} \and
Ke Wang}

\authorrunning{E. Hoseinzade and K.Wang}

\institute{Simon Fraser University, Burnaby, Canada \\
\email{\{ehoseinz,wangk\}@sfu.ca}}

\maketitle              

\begin{abstract}
This study addresses the challenge of detecting semantic column types in relational tables, a key task in many real-world applications. While language models like BERT have improved prediction accuracy, their token input constraints limit the simultaneous processing of intra-table and inter-table information. We propose a novel approach using Graph Neural Networks (GNNs) to model intra-table dependencies, allowing language models to focus on inter-table information. Our proposed method not only outperforms existing state-of-the-art algorithms but also offers novel insights into the utility and functionality of various GNN types for semantic type detection. The code is available at https://github.com/hoseinzadeehsan/GAIT

\end{abstract}

\section{Introduction}
Accurately identifying (or tagging) the semantic types of columns inside a table is crucial for different information retrieval tasks like data cleaning \cite{limaye2010annotating}, schema matching \cite{rahm2001survey}, and data discovery \cite{fernandez2018aurum}. One emerging application is automatically tagging sensitive columns in a table, such as personal information, before 
deciding what information can be released. 
Previous works showed that machine learning approaches outperform traditional methods in predicting semantic types \cite{hul2019sherlock,zhang2019sato,chen2019colnet,chen2019learning}. Sherlock \cite{hul2019sherlock}, a single-column prediction framework, feeds various features of a column to a deep feed-forward neural network to get the prediction. This method ignores the global context and the dependencies between columns, making it difficult to distinguish the semantic types in cases like in Figure \ref{fig: problem}. SATO \cite{zhang2019sato} improves upon Sherlock by adding a topic modeling module and a structured prediction module on top of Sherlock, to jointly predict semantic types of all the columns in a table by leveraging the topic of a table and the dependencies between columns in a table.

Building on the trend of applying machine learning to tabular data, researchers have started using language models like BERT \cite{devlin2018bert}. By feeding tables to BERT, which was originally designed for textual data, they exploit its extensive pre-training. This adaptation has created new frameworks that fine-tune BERT for column type annotation \cite{deng2022turl}. In addition to the values of the target column, two other sources of information can be used to improve the accuracy of semantic type annotation: intra-table information refers to other columns in the same table and inter-table information refers to other tables in the data. Given that language models have a small limit on the number of input tokens (BERT takes a maximum of 512 tokens), column type annotation models are developed to handle only one of the two mentioned sources of information.

Incorporating intra-table information has led to multi-column prediction approaches \cite{yin2020tabert} that are designed to address the limitation of single-column prediction models in situations like Fig \ref{fig: problem} by accounting for broader table context, specifically column relationships and information. TABBIE \cite{iida2021tabbie} encodes rows and columns of a table respectively to get a better understanding of tables. The most prominent work in this category is Doduo \cite{suhara2022annotating} where BERT is modified to receive the whole columns of a table and predict their semantic types together.

Having inter-table information \cite{wang2021tcn} can be a huge help in cases where the target column does not have enough high-quality data to make a good semantic prediction. For example, if a table has a column with entries like 'Orange' and 'Peach', the semantic type is ambiguous. However, by identifying and augmenting this column with columns of similar tables that have entries like 'Red' and 'Blue', the semantic type becomes clearer, indicating that this column is likely about colors rather than fruits. The most recent work, RECA \cite{sun2023reca}, in addition to the values of the target column, identifies values of the most useful similar tables and feeds them to BERT to get the semantic type of the target column.

However, due to the small limit on the number of input tokens of language models like BERT, Doduo and RECA have the following drawbacks:

\begin{figure*}[t]

\centering
\includegraphics[width=0.8\textwidth]{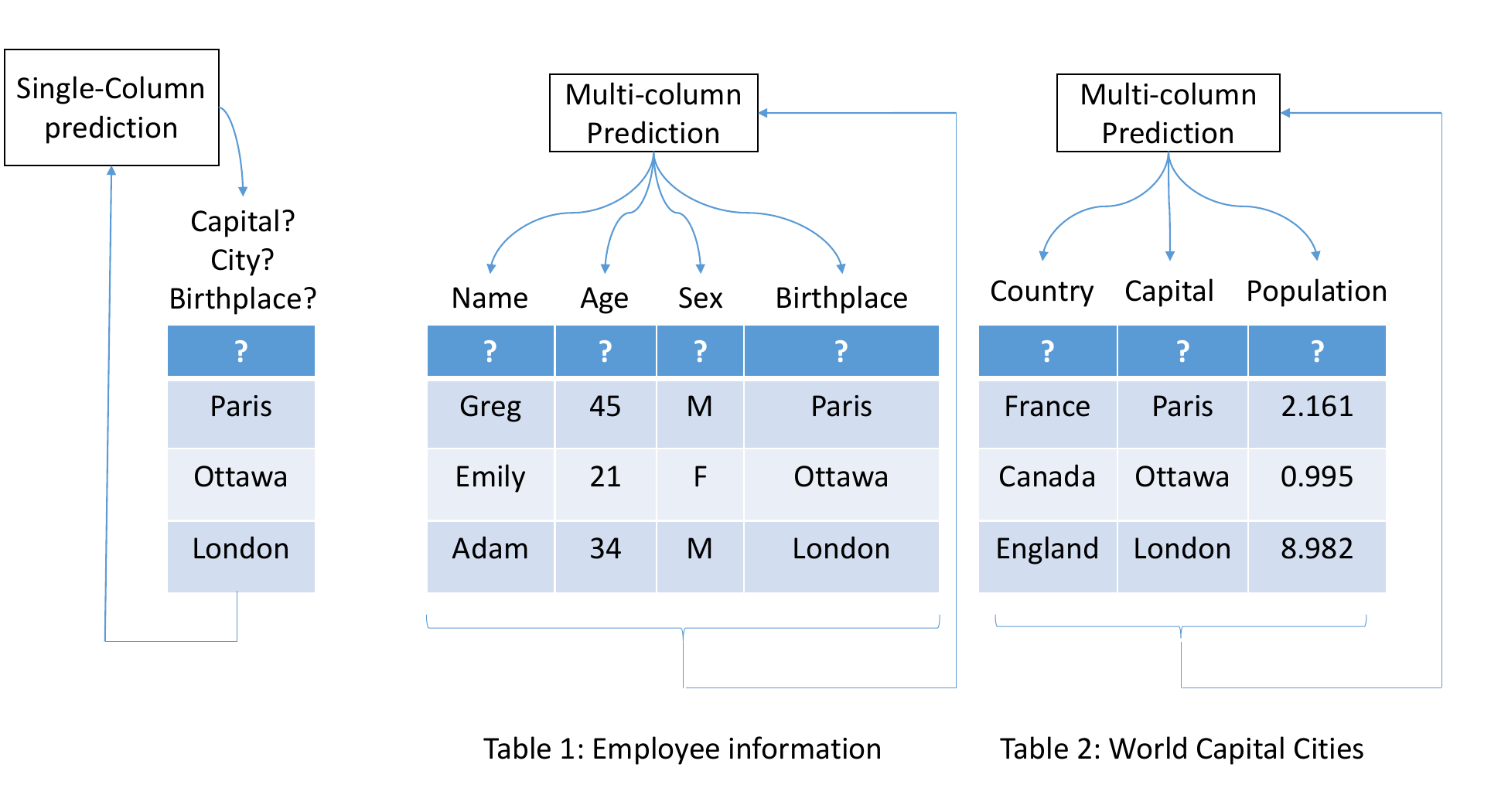} 

\caption{The two tables on the right both have the column containing the values ``Paris", ``Ottawa" and ``London". Without considering information coming from other columns it is difficult for a single-column prediction model to detect the actual semantic types of these columns. The multi-column prediction will label these columns correctly by jointly predicting all columns in a table.}
\label{fig: problem}
\vspace{-5mm}
\end{figure*}

\begin{enumerate}
    \item Doduo feeds the whole table to BERT and because of that, it poses difficulties in handling wide tables. For instance, the average number of columns of tables in Open Data is 16, but there's a large variance with some tables having hundreds of columns. Furthermore, Doduo is not designed in a way that can incorporate inter-table context information and ignores this useful source of information \cite{sun2023reca}.

    \item RECA incorporates inter-table but not intra-table information, predicting the semantic type of each column individually. This approach enables RECA to handle wide tables within the language model's small token limit, as it doesn't need to model the entire table at once. However, this means it overlooks the valuable information in column relationships, crucial in complex scenarios like Fig \ref{fig: problem}, where semantic types are difficult to distinguish.

\end{enumerate}

Thus, a question is whether it is possible to incorporate both inter-table and intra-table information without suffering from the difficulty of handling wide tables as in Doduo and to benefit from the generalization power of the language model based approach. Inspired by recent developments in computer vision such as visual reasoning, object detection, and scene graph generation \cite{chen2018iterative,li2017scene}
where a main task is tagging the objects inside an image by leveraging the relations among the objects in the image, we propose to augment any single-column prediction framework, especially those incorporating inter-table information like RECA (addressing the drawback of Doduo) by a graph neural network (GNN) module to model the whole dependencies between columns. Thus, our framework incorporates both inter-table and intra-table information. In particular, we model each table by a graph with the nodes representing the columns and the edges representing the dependencies between columns through Message Passing of GNN. By considering the dependencies between all pairs of columns, this approach, called GAIT (\textbf{G}raph b\textbf{A}sed semant\textbf{I}c \textbf{T}ype detection), addresses the above drawback of RECA, making it a multi-column prediction framework. The challenge is how to represent the features of a column by a node so that Message Passing can leverage the dependencies among columns. 

GAIT stands out in efficiently handling wide tables and benefiting from a language model based approach by building on top of models like RECA that are single-column predictions and language model based. While Doduo's effectiveness decreases in scenarios with minimal column dependency, and RECA faces challenges when similar tables are scarce, GAIT's integration of both inter-table and intra-table information makes it a competitive model in these diverse scenarios. This dual-data approach enables GAIT to maintain its performance and effectively address the limitations encountered by models focusing on either inter-table or intra-table information alone.

\section{Related works}

Column type prediction methods are typically grounded into two categories, i.e., deep learning based frameworks and language model based frameworks.

\textbf{Deep learning based models}.
ColNet \cite{chen2019colnet} uses DBpedia cell value lookups to create examples and trains a CNN with Word2Vec embeddings. HNN \cite{chen2019learning}, models intra-column semantics, enhancing Colnet. Sherlock \cite{hul2019sherlock} employs column statistics, paragraph, word, and character embeddings to predict column types through a neural network. SATO \cite{zhang2019sato} builds on Sherlock, adding topic modeling features and adjusting predictions for column dependencies using a CRF.

\textbf{Language model based models}.
Language models, like BERT \cite{devlin2018bert}, have been adopted for table tasks \cite{wang2021tuta} including column type prediction. TaBERT \cite{yin2020tabert} utilizes BERT as a base model to capture the table content features. TURL \cite{deng2022turl} is pre-trained unsupervisedly, using a visibility matrix for row and column context, and then fine-tuned for table-related tasks. 
TABBIE \cite{iida2021tabbie} separately processes the rows and columns of tables to give a better understanding of them. Doduo \cite{suhara2022annotating} predicts all the columns of a table together by feeding the whole table to BERT. RECA \cite{sun2023reca} incorporates inter-table context information by finding and aligning relevant tables. TCN \cite{wang2021tcn} suggests using both intra-table and inter-table information for column type prediction. However, it needs table schema and page topic, which many datasets, like Webtables and Semtab don't have.

\textbf{Summary:} Most of the previous works \cite{yin2020tabert,suhara2022annotating,deng2022turl,iida2021tabbie}, except for TCN and  RECA, do not incorporate inter-table information for prediction. TCN \cite{wang2021tcn} requires having table schema and page topic, which does not exist in many datasets. RECA does not incorporate intra-table dependencies.
Our GAIT predicts semantic types of columns exclusively based on the content of the tables by integrating both intra-table and inter-table information.

\section{Problem Definition}

We aim to predict the semantic types of the columns of a given table with missing column headings. This problem is called \textit{table annotation}. 
To learn to predict semantic types, a collection of labeled tables is given as the training data $D$, where each table $t(c_1, c_2, ..., c_n)$ consists of $n$ columns and each column is labeled as one of the $k$ pre-defined semantic types, also called \textit{classes}, e.g., Age, Name, Country (note that semantic types are different from atomic types like integer and string). Note that the number of columns $n$ and rows can differ for different tables. 
Typically, the first step is to extract a feature vector (embedding) to represent a column $c_i$. 
After applying a feature extractor function $\phi$ to the values of a column $c_i$ and potential inter-table information related to $c_i$, an $m$-dimension feature (embedding) vector $\psi_i$ is generated for column $c_i$. 
The rest of the task is to learn a mapping $f$ that, given $\psi=<\psi_1, ..., \psi_n>$ of a table of $n$ unlabeled columns, predicts the classes 
for the $n$ columns in the table. 

\begin{figure}[t]
\centering
\includegraphics[width=0.9\columnwidth]{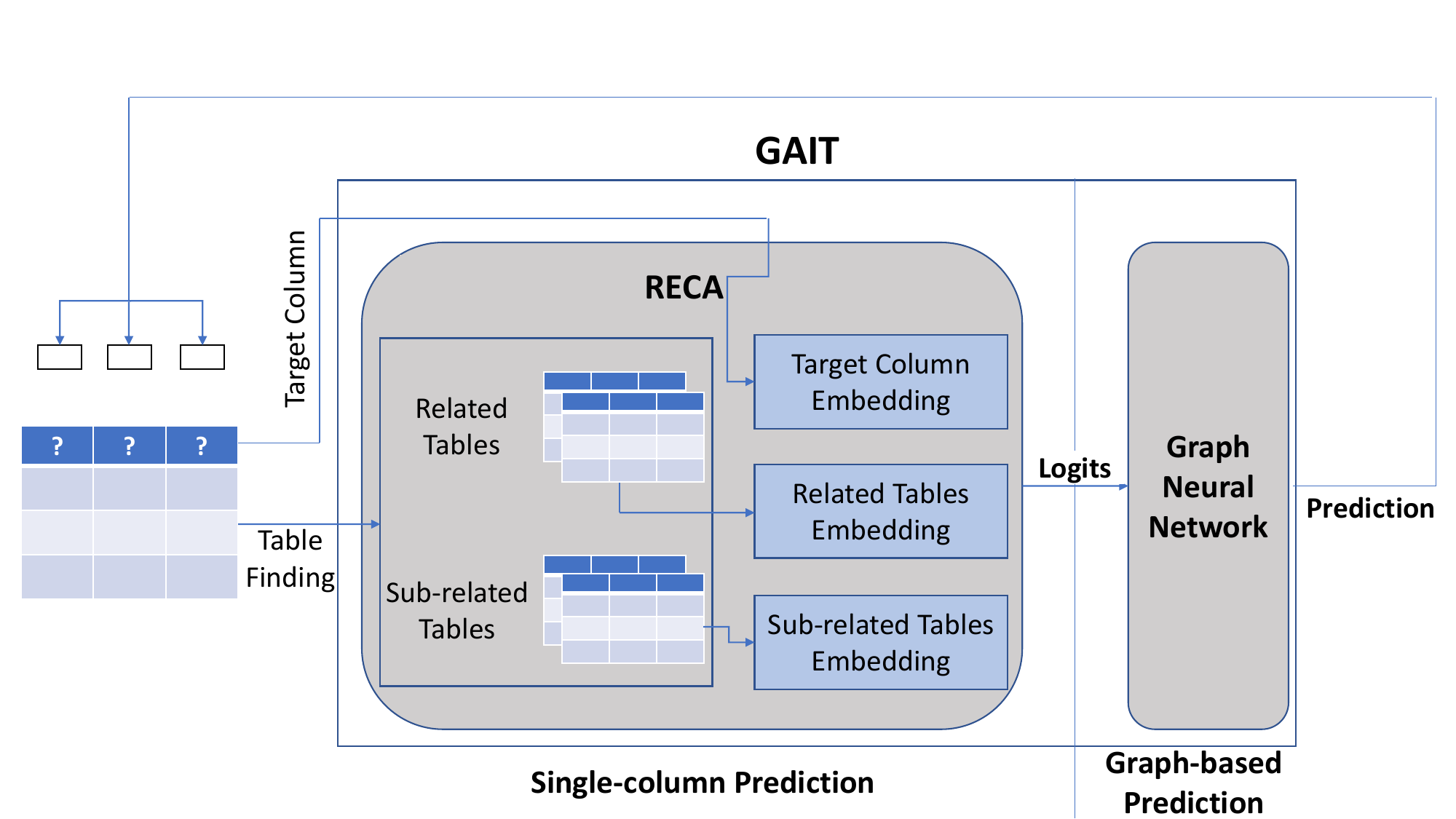} 
\caption{The framework of GAIT: GAIT adds a GNN learning on top of the single-column prediction module, which is RECA in this work. The output of RECA is a class distribution for each column in a table, which provides the initial hidden state of the node representing that column in the GNN. For a table with $n$ columns, RECA is performed $n$ times. Then, the GNN learns the best representations of the hidden states of all nodes to minimize a loss function, through Message Passing that models the dependencies between columns.}
\label{fig: overall gait}
\end{figure}

\section{GAIT}
Figure \ref{fig: overall gait} shows the framework of GAIT. It is built on a single-column prediction 
provided by RECA. 
We opted for RECA due to its high performance in column type annotation as a result of incorporating useful inter-table information from other relevant tables.
That said, GAIT's design is versatile. While we utilize RECA, other single-column prediction modules can be integrated.
GAIT employs a GNN in which each graph represents a table with the nodes representing the columns in the table. The initial representation of each node is the logits outputted by RECA 
for the represented column. Once fed with such class distributions as input, the training of GNN is responsible for capturing the dependencies of classes among columns and does not further involve the lower-level single-column prediction module. This approach treats the preliminary prediction of the single-column prediction of RECA as the node features for training GNN, which is more efficient than concatenating the networks of low-level modules into a giant neural network. Our method stacks a GNN 
as a meta-learner on top of RECA (i.e., two classifiers) instead of concatenating RECA and GNN into one classifier, which improves the overall performance according to 
stacked generalization technique \cite{wolpert1992stacked}. 
We now present more details.

\subsection{Single-column Prediction}
The single-column prediction is responsible for generating the preliminary prediction of each column. We use RECA \cite{sun2023reca} for this task. In the RECA process, the primary goal is to improve the understanding of a target column in the main table by integrating relevant data from other tables. The process begins by identifying named entities across all tables. Each entity is assigned a type from a predefined set, with the most common type within a column being selected as its representative named entity type. Following this, RECA constructs the named entity schema for each table, which includes the named entity types of all its columns. The next step is to find the topical relevance of other tables to the main table. This is done by calculating the Jaccard similarity between the words in the main table and other tables. Tables that are similar enough are chosen as candidate tables for further analysis. Among these candidates, tables with the same named entity schema as the main table are labeled as relevant tables. Additionally, tables with similar, but not identical, named entity schemas are called sub-related tables. The final step involves combining the data from the target column in the main table with data from columns having the same named entity type in both related and sub-related tables and feeding it to a language model, BERT, to find the semantic type of the target column.

\subsection{Graph-based Prediction}

\textbf{Graph Modeling of a Table} \label{modeling}
 The training data for GNN is a collection of graphs organized into several mini-batches, where each graph corresponds to a table in the original training data. For a table with $n$ columns, we create a graph of $n$ nodes where each node represents a column in the table and create an edge between each pair of columns. Initially, each node $u$ of a graph has the representation $h^0_u$ initialized to the logits $<o_1,...,o_k>$ outputted by RECA for the corresponding column, which has one value for each class. This initial state represents the class bias of single-column prediction. In addition, each node is associated with the true class of the represented column.

\textbf{Message Passing} \label{message}
Subsequently, the representation of all the nodes in a mini-batch of graphs is updated through the Message Passing mechanism of GNN along edges. For this purpose, we consider three different types of GNNs, graph convolutional network (\textbf{GCN}) \cite{kipf2016semi}, gated graph neural network (\textbf{GGNN}) \cite{li2015gated}, and graph attention network (\textbf{GAT}) \cite{velivckovic2017graph}, with the following $UPDATE$ functions where $\sigma$ is the activation function, $N(u)$ is a list of nodes connected to node $u$, $h_u^{s}$ is the representation (also called \textit{embedding}) of node $u$ at step $s\geq 0$, $W^{(s)}$ is a model parameter: 
\begin{itemize}
    \item \textbf{GCN:} assigns equal weights to all the neighbor nodes while updating the embedding of each node (Eq \ref{eq: GCN}).

    \begin{equation}
        h_{u}^{(s+1)} = \sigma ( \sum_{v \in N(u)\cup u} \frac{W^{(s)}_{} h_v^{(s)}}{\sqrt{|N(u)| |N(v)|}})
    \label{eq: GCN}
    \end{equation}
    
    \item \textbf{GGNN:} uses gated recurrent unit (GRU) to evaluate messages coming from adjacent nodes while updating the embedding of each node (Eq \ref{eq: ggnn}).
    \begin{equation}
        h_u^{s+1} = GRU(h_u^s, \sum_{v \in N(u)} W^{(s)}_{} h_v^s)
    \label{eq: ggnn}
    \end{equation}

    \item \textbf{GAT:} updates node embedding (Eq \ref{eq: gat1_multi}) according to the multi-head attention weights (Eq \ref{Eq: attention_multi}), where $K$ is the number of attention heads, $a^{(s,k)}$ and $W^{(s,k)}$ are model parameters for attention head $k$, and $\oplus$ is concatenation.
    \begin{equation}
        h_u^{(s+1)} = \oplus_{k=1}^{K} (\sigma \sum_{v \in N(u)\cup\{u\}} \alpha^{(s,k)}_{u, v} W^{(s,k)} h_v^{(s)}) 
    \label{eq: gat1_multi}
    \end{equation}

    \begin{equation}
        \alpha^{s,k}_{u, v} = \frac{\exp ( ReLU ( a^{(s,k)^T} ( W^{(s,k)} h_u^{(s)} \oplus W^{(s,k)} h_v^{(s)} ) ) )}{\sum_{v' \in N(u)\cup\{u\}} \exp ( ReLU ( a^{(s,k)^T} ( W^{(s,k)} h_u^{(s)} \oplus W^{(s,k)} h_{v'}^{(s)} ) ) )}
    \label{Eq: attention_multi}
    \end{equation}

\end{itemize}

The $UPDATE$ function is applied to each node $u$ in the mini-batch of graphs for $S$ steps, where $S$ is the number of hidden layers and output layer. $h^S_u$ has one unit for each of the $k$ classes and serves the final output for the $k$ classes. The class prediction for the node $u$ is given by applying softmax to $h^S_u$. 

The main difference among GCN, GGNN, and GAT is in their treatment of adjacent nodes (columns). GAT uses an attention mechanism to assign varying weights to these nodes based on their importance. In contrast, GCN averages the features of neighbor nodes, while GGNN processes these features through a GRU to determine their relevance before updating the node embeddings. 

\textbf{Loss Function} \label{loss} 
 Given the logit vector $h^S_u$ for a node $u$  with the true class $class_u$, the \textit{loss} for this node is computed by the negative log-likelihood. The loss for a mini-batch of graphs is the sum of the loss of all the nodes inside the mini-batch of graphs (Eq \ref{eqation: loss}).
 We update the model parameters 
 to minimize the loss of a mini-batch by performing stochastic gradient descent.

\begin{equation}
    loss = \sum_{u=1}^{\#node} - \log(\frac{\exp(h_u^{S}[class_u])}{\sum_{m=1}^{k} \exp(h_u^{S}[m])})
    \label{eqation: loss}
\end{equation}

\subsection{Overall Prediction}
After training the GNN, to classify columns in a new table $t$, we first get the output of RECA (before softmax) for each column in $t$. These outputs are the initial representation $h^0_u$ of nodes $u$ in the graph representing the table $t$. Then, message passing is done using the learned parameters in the training phase to get the predicted class for each node, which is the predicted class for the corresponding column in $t$. Fig \ref{fig: overall gait} shows how GAIT predicts labels of a table.

\section{Evaluation}

\subsection{Evaluation Method}

\textbf{Performance metrics}.
Like previous works  \cite{hul2019sherlock,zhang2019sato,suhara2022annotating,sun2023reca},
we collect \textit{weighted f-score}  and \textit{macro f-score} on the test data. The former is the average of f-score of all classes, weighted by class frequencies, and the latter is the average of treating all classes equally, regardless of their frequencies. The macro f-score better reflects the model performance on infrequent classes. We evaluate model performance using 5-fold cross-validation, reporting the mean and standard deviation of the above f-scores from the test split of each fold.

\textbf{Datasets}. 
We use two datasets summarized in Table \ref{table: big table statistic}. 

Webtables \cite{zhang2019sato,suhara2022annotating,sun2023reca}: This dataset contains 32262 tables and 78 unique classes extracted from the Webtables directory of VizNet \cite{hu2019viznet}. We use \textit{exactly the same} 5-fold cross-validation split as in \cite{sun2023reca}
, which splits the tables (not columns) into a train set and test set in 5-folds. So, we copy directly the f-scores of the baseline algorithms (more details on the baselines below) except for RECA from \cite{sun2023reca}.

Semtab2019 \cite{sun2023reca}: It contains 3045 tables and 275 unique classes. While this dataset covers wider tables (an average of 4.5 columns per table), only 7603 columns are annotated. The split proposed in RECA \cite{sun2023reca} randomly divided columns (not tables) into train, validation, and test sets. Although the column-wise splitting of data makes sense for RECA due to the column-wise prediction of RECA, GAIT requires having a full table to model the dependencies between columns in the table. Therefore, our 5-fold validation splits tables (instead of columns) into the train set and test set for this dataset. At each fold, we further split the train set into 80\% for training and 20\% for validation.

\begin{table}[t]
\caption{Datasets used and the number of tables with the specified number of columns.}
\centering
\setlength{\tabcolsep}{8pt} 
\resizebox{0.7\columnwidth}{!}{%
\scriptsize 
\begin{tabular}{ccccc}
    \toprule
     Dataset & \#types & \#tables & \#Col & avg col \\ 
    \midrule
     Semtab & 275 & 3045 & 7603 & 4.5 \\
     Webtable & 78 & 32262 & 74141 & 2.3 \\
    \bottomrule
\end{tabular}%
}
\label{table: big table statistic}
\end{table}

\textbf{Algorithms for comparison}. All experiments were conducted with Tesla V100s. We used the publicly available source code of RECA\footnote{https://github.com/ysunbp/RECA-paper} for the single-column prediction module of GAIT. The GNN module of GAIT was implemented using the deep graph library \cite{wang2019deep}
, Adam optimization  
with a learning rate of $1e-3$ and weight decay of $5e-4$ for training. We trained GCN, GGNN, and GAT for 100, 200, and 100 epochs respectively. To optimize the GAT structure, we tested various \# attention heads ([1, 2, 4, 8, 12]) and update steps $S$ ([1, 2, 3, 4]), selecting the best model from the validation set as default. Similarly, for GGNN and GCN, we determined the default model by experimenting with different update steps $S$ ([1, 2, 3, 4]). Three algorithms for GAIT were finally chosen: $\textnormal{\textbf{GAIT}}_{\textnormal{\textbf{GAT}}}$  (GAT with $S=2$), \textbf{GAIT\textsubscript{GGNN}} (GGNN with $S=3$), and \textbf{GAIT\textsubscript{GCN}} (GCN with $S=2$).

Since GAIT incorporates RECA as its single-column prediction module, naturally we evaluate GAIT against the baseline methods outlined in RECA's paper and RECA itself. 
These baselines are described below and their source codes are publicly available and are used for our evaluation:

\begin{itemize}
    \item Sherlock \cite{hul2019sherlock}: Sherlock is a deep learning model that extracts character-level, word-level, paragraph-level and global-level statistical features from tables to form vector representations for table columns. 
     \item TaBERT \cite{yin2020tabert}: 
     TaBERT simultaneously analyzes queries and a table, selecting three crucial rows to create table content snapshots. It then uses BERT to develop representations for each table column, aiding in classification.
     \item TABBIE \cite{iida2021tabbie}: improves TaBERT by separately processing the rows and columns of tables. The embedding of the target column is used for prediction.
     \item Doduo \cite{suhara2022annotating}: 
     Modifies BERT to feed the whole columns of a table to BERT and predicts the semantic types of all of the columns in a table together.
     \item RECA \cite{sun2023reca}: RECA finds relevant tables for the target table and uses the information coming from these tables and the values of the target column to predict the semantic type of the target column. 
\end{itemize}
We do not compare with SATO \cite{zhang2019sato} and TURL \cite{deng2022turl} as Doduo outperformed them. Since TCN \cite{wang2021tcn} requires having table schema and page topic \cite{sun2023reca}, it cannot be applied to our datasets. 

\subsection{Results}
Table \ref{table:transposed_results} Shows the performance of GAIT and the baseline algorithms. GAIT outperforms Sherlock by a large margin. The main reason behind the poor performance of Sherlock compared with other models is its simplicity. While other models including GAIT utilize language models for semantic type prediction, Sherlock relies on simple semantic features to do so. Furthermore, Sherlock does not use intra-table or inter-table information for prediction.

\begin{table}[t!]
 \caption{Macro f-score and weighted f-score.}
 \centering
 \setlength{\tabcolsep}{5pt} 
 \begin{tabular}{lcccc} 
 \toprule
  & \multicolumn{2}{c}{Semtab} & \multicolumn{2}{c}{Webtables} \\
 \cmidrule(lr){2-3} \cmidrule(lr){4-5}
 Model & Weighted f-score & Macro f-score & Weighted f-score & Macro f-score \\
 \midrule
 sherlock \cite{hul2019sherlock} & 0.638$\pm$0.009 & 0.417$\pm$0.017 & 0.844$\pm$0.001 & 0.670$\pm$0.010 \\
 TaBERT \cite{yin2020tabert} & 0.756$\pm$0.011 & 0.401$\pm$0.025 & 0.896$\pm$0.005 & 0.650$\pm$0.011 \\
 TABBIE \cite{iida2021tabbie} & 0.798$\pm$0.012 & 0.542$\pm$0.022 & 0.929$\pm$0.003 & 0.734$\pm$0.019 \\
 Doduo \cite{suhara2022annotating} & 0.819$\pm$0.010 & 0.565$\pm$0.021 & 0.928$\pm$0.001 & 0.742$\pm$0.012 \\
 RECA \cite{sun2023reca} & 0.825$\pm$0.015 & 0.583$\pm$0.019 & 0.935$\pm$0.032 & 0.783$\pm$0.017 \\
 \midrule
$\textnormal{GAIT}_{\textnormal{GGNN}}$ & 0.844$\pm$0.003 & 0.606$\pm$0.018 & 0.936$\pm$0.003 & 0.797$\pm$0.022 \\
$\textnormal{GAIT}_{\textnormal{GCN}}$ & 0.845$\pm$0.006 & 0.622$\pm$0.020 & 0.939$\pm$0.004 & 0.794$\pm$0.017 \\
$\textnormal{GAIT}_{\textnormal{GAT}}$ & \textbf{0.852$\pm$0.004} & \textbf{0.643$\pm$0.017} & \textbf{0.940$\pm$0.003} & \textbf{0.799$\pm$0.019}  \\
 \bottomrule
 \end{tabular}
 \label{table:transposed_results}
\end{table} 

Among language model based models TaBERT shows the worst performance because it was initially developed for table semantic parsing and column embeddings generated
by TaBERT are not suitable for column type annotation \cite{sun2023reca}. RECA, single-column prediction module of GAIT, outperforms both TABBIE and Doduo. TABBIE and Doduo use the limited input tokens of language models to process intra-table context while ignoring the inter-table context information when generating the embeddings of the target columns. However, RECA mainly focuses on extracting useful inter-table context information to enhance the embeddings of the target columns \cite{sun2023reca}. 

GAIT with different GNNs outperforms RECA, and by extension TABBIE and Doduo in both datasets. In particular, GAIT shows about 6\% and 2.7\% improvement in macro and weighted f-scores over RECA in the Semtab-dataset. These results prove that modeling the dependencies between columns in a table, which is the main advantage of GAIT over RECA, is useful. GAIT successfully applies a GNN on top of RECA to do so. Among different variations of GAIT, GAT shows the best performance. Assigning different weights to adjacent nodes (columns) according to their importance when updating representation of a node is the key to the superior performance of GAT compared to GCN and GGNN. 

In both datasets, GAIT's improvement is larger on the macro f-score than the weighted f-score. This means that infrequent classes that label fewer columns benefit more from the whole dependency approach of the GNN approach. Such classes have less presence in the data and their learning tends to rely on the dependencies on other columns in a table. GAIT provides a mechanism to leverage such dependencies. This also explains why GAIT shows a better enhancement in the Semtab dataset compared to Webtables dataset. The 3045 tables of Semtab have 275 semantic types for columns while the 32262 tables of Webtables are limited to 78 semantic types. Consequently, many more infrequent classes in Semtab can benefit from modeling the whole dependencies of GAIT. 

\begin{figure*}[t]
\centering
\vspace{-15mm}
\includegraphics[width=0.9\textwidth]{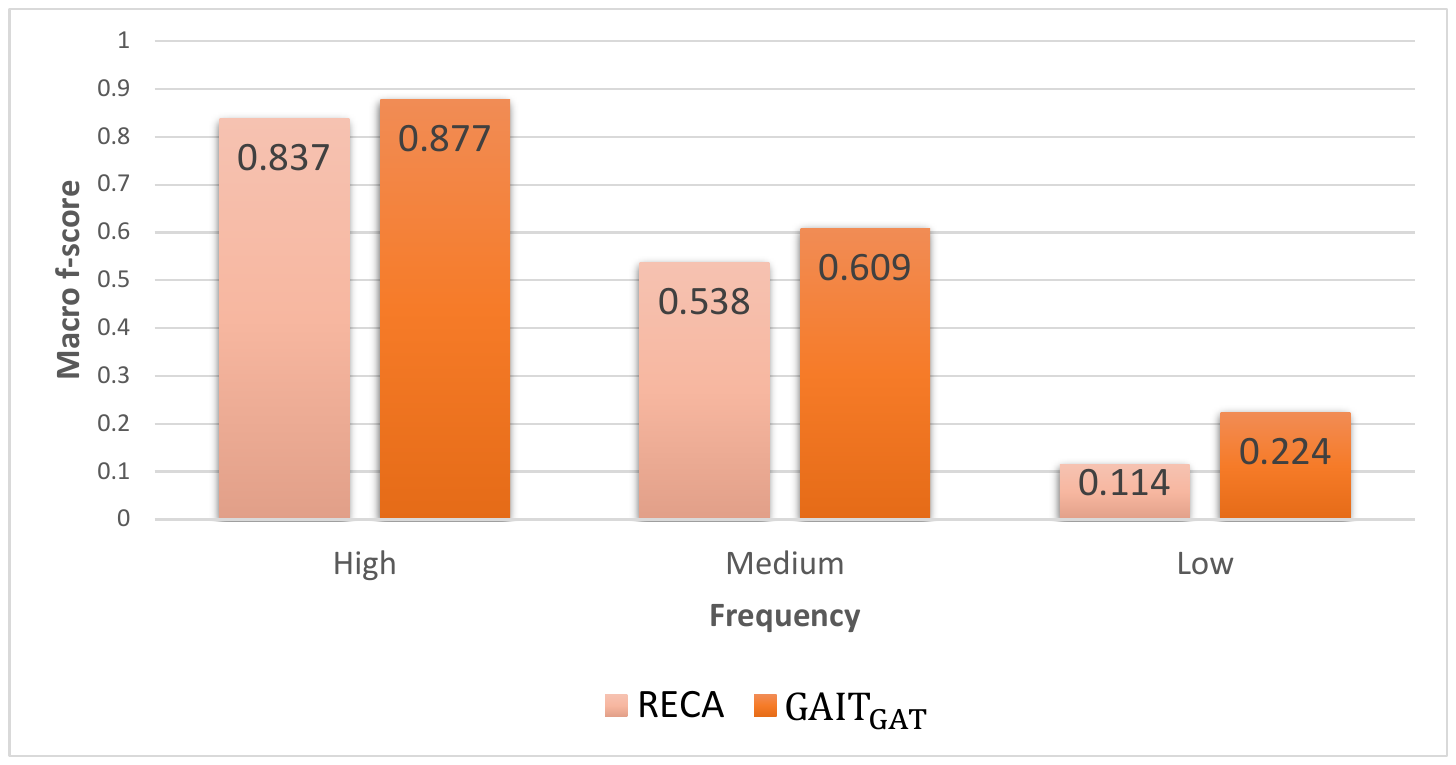}
\vspace{-15mm}
\caption{The macro f-score of $\textnormal{GAIT}_{\textnormal{GAT}}$ and RECA on Semtab dataset, for High, Medium, and Low-frequency classes.} 
\label{fig: frequency}
\end{figure*}

To provide a better insight into this improvement, we divide the 275 classes of Semtab dataset into three equally sized bins of High, Medium, and Low frequencies (about 92 classes in each bin) according to the columns labeled by classes and show the macro f-score of the classes in each bin for $\textnormal{GAIT}_{\textnormal{GAT}}$ (best GAIT) and RECA (best baseline) in Fig \ref{fig: frequency}. While $\textnormal{GAIT}_{\textnormal{GAT}}$ improves RECA in all the three bins, the bigger improvements happen in the low-frequency bins, for example, the absolute improvement of 11\% or the relative improvement of 96.5\% for the Low bin. 
Fig \ref{fig: frequency} also demonstrates that the real challenge in developing column type annotation models is how to have a reliable prediction for medium and low-frequency classes as the performance for high-frequency classes is already good enough. The large improvements of $\textnormal{GAIT}_{\textnormal{GAT}}$ over RECA in low-frequency classes is a clear sign of its superiority in handling such classes.

We also study the impact of the number of columns in a table on both $\textnormal{GAIT}_{\textnormal{GAT}}$ and RECA. Table \ref{table: results table size} shows the improvement of $\textnormal{GAIT}_{\textnormal{GAT}}$ on macro and weighted f-scores over RECA \textit{separately} for tables of a different number of columns. As the number of columns in a table increases, the performance of RECA, which is also the single-column prediction module of GAIT, increases. Having more columns in a table better reveals context of that table, so RECA can find more relevant inter-table information which is beneficial to both RECA and GAIT.
Thus, the need for dependencies between columns in $\textnormal{GAIT}_{\textnormal{GAT}}$ decreases. 
The column dependency method GAIT improves RECA mainly for low-frequency classes and tables of 2 to 4 columns, as in case of Semtab.

\begin{table}
\caption{The f-score improvement of $\textnormal{GAIT}_{\textnormal{GAT}}$ over RECA by tables of different number of columns.}
\centering
\setlength{\tabcolsep}{1pt} 
\resizebox{\textwidth}{!}{%
\normalsize 
\begin{tabular}{ccccccccc}
    \toprule
    \multicolumn{1}{c}{} & \multicolumn{4}{c}{Semtab} & \multicolumn{4}{c}{Webtables} \\
    \cmidrule(lr){2-5} \cmidrule(lr){6-9}
    & \multicolumn{2}{c}{macro f-score} & \multicolumn{2}{c}{weighted f-score} & \multicolumn{2}{c}{macro f-score} & \multicolumn{2}{c}{weighted f-score} \\
    \cmidrule(lr){2-3} \cmidrule(lr){4-5} \cmidrule(lr){6-7} \cmidrule(lr){8-9}
    \#col & RECA & $\textnormal{GAIT}_{\textnormal{GAT}}$ & RECA & $\textnormal{GAIT}_{\textnormal{GAT}}$ & RECA & $\textnormal{GAIT}_{\textnormal{GAT}}$ & RECA & $\textnormal{GAIT}_{\textnormal{GAT}}$ \\ 
    \midrule
    2 & 0.563 & 0.603 (+4.0\%) & 0.798 & 0.828 (+3.0\%) & 0.738 & 0.758 (+2.0\%) & 0.932 & 0.936 (+0.4\%) \\ 
    3 & 0.545 & 0.616 (+7.1\%) & 0.797 & 0.827 (+3.0\%) & 0.743 & 0.762 (+1.9\%) & 0.927 & 0.930 (+0.3\%) \\
    4 & 0.566 & 0.618 (+5.2\%) & 0.865 & 0.880 (+1.5\%) & 0.727 & 0.746 (+1.9\%) & 0.960 & 0.961 (+0.1\%) \\
    5 & 0.664 & 0.682 (+1.8\%) & 0.862 & 0.862 (+0.0\%) & 0.540 & 0.548 (+0.8\%) & 0.978 & 0.978 (+0.0\%) \\
    \bottomrule
\end{tabular}%
}
\label{table: results table size}
\end{table}

\vspace{-1mm}
\section{Conclusion}
\vspace{-1mm}
Language model based approaches recently showed promising results in column type annotation thanks to the semantic knowledge preserved in them. This paper addresses some drawbacks of previous language model-based approaches, namely, failing to incorporate inter-table and intra-table information simultaneously due to the input token limit of language models. Our solutions, GAIT, employ graph neural networks to model the intra-table dependencies, letting language models focus on handling inter-table information. Experiments on different datasets provide evidence of the effectiveness of our solutions. Looking ahead, considering the recent advancements in large language models (LLMs) for column type annotation \cite{feuer2023archetype,korini2023column,zhang2023jellyfish,li2023table} exploring alternative LLMs beyond BERT to address inter-table information could be a promising future research. \\

\textbf{Acknowledgement.} The work of Ke Wang is supported in part by a discovery grant from Natural Sciences and Engineering Research Council of Canada.

\bibliographystyle{splncs04}
\bibliography{references}
\end{document}